\documentclass[10pt,twocolumn,letterpaper]{article}

\usepackage{cvpr} 
\usepackage{amsmath}
\usepackage{amssymb}
\usepackage{booktabs}
\usepackage{pgfplots}
\usepackage{graphicx} 
\usepackage{tikz}
\usetikzlibrary{matrix, positioning, fit}
\usetikzlibrary{decorations.pathmorphing}
\usepackage{colortbl}   
\usepackage{pgfplots}   
\usepackage{adjustbox}  
\usepackage{caption}    
\usepackage{multirow}
\usepackage{url}
\usepackage{longtable}
\usepackage{makecell}
\usepackage[hang,flushmargin]{footmisc}

\newcommand{\cellbarsgan}[2]{%
  \begin{tikzpicture}

    \ifdim #1 pt > 0pt
      \pgfmathsetmacro{\barwidth}{min(#1*0.05, 1)}
      \fill[green!50] (0,0) rectangle (\barwidth,0.35);
    \else
      \pgfmathsetmacro{\barwidth}{max(#1*0.05, -1)}
      \fill[red!50] (1+\barwidth,0) rectangle (1,0.35); 
    \fi
    
    \node[anchor=center] at (0.5,0.175) {#2}; 
  \end{tikzpicture}%
}

\newcommand{\cellbardif}[2]{%
  \begin{tikzpicture}
    \ifdim #1 pt > 0pt
      \pgfmathsetmacro{\barwidth}{min(#1, 1)}
      \fill[green!50] (0,0) rectangle (#1,0.35);
    \else
      \pgfmathsetmacro{\barwidth}{max(#1, -1)}
      \fill[red!50] (1+\barwidth,0) rectangle (1,0.35);
    \fi
    \node[anchor=center] at (0.5,0.175) {#2}; 
  \end{tikzpicture}%
}

\usepackage{xcolor}

\definecolor{ss}{RGB}{142,142,142}
\definecolor{gr0}{RGB}{13,13,13}
\definecolor{gr1}{RGB}{185,152,49}

\pgfdeclarelayer{background}
\pgfsetlayers{background,main}

\pgfmathdeclarefunction{gauss}{3}{%
  \pgfmathparse{1/(#3*sqrt(2*pi))*exp(-0.5*((x-#2)/#3)^2)}%
}

\pgfmathdeclarefunction{gauss_gauss}{5}{%
  \pgfmathparse{0.85/(#3*sqrt(2*pi))*exp(-0.5*((x-#2)/#3)^2)+0.85/(#5*sqrt(2*pi))*exp(-0.5*((x-#4)/#5)^2)}%
}

\begin{document}

\title{Bias Analysis for Synthetic Face Detection: A Case Study of the Impact of Facial Attributes}

\author{
Asmae Lamsaf$^{1,2}$ \quad
Lucia Cascone$^3$ \quad
Hugo Proença$^1$ \quad
João Neves$^{1,2}$ \\
\small $^1$University of Beira Interior, Portugal \quad 
\small $^2$NOVA LINCS \quad
\small $^3$University of Salerno, Italy
}

\maketitle
\thispagestyle{empty}

\begin{abstract}
Bias analysis for synthetic face detection is bound to become a critical topic in the coming years. Although many detection models have been developed and several datasets have been released to reliably identify synthetic content, one crucial aspect has been largely overlooked: these models and training datasets can be biased, leading to failures in detection for certain demographic groups and raising significant social, legal, and ethical issues. In this work, we introduce an evaluation framework to contribute to the analysis of bias of synthetic face detectors with respect to several facial attributes. This framework exploits synthetic data generation, with evenly distributed attribute labels, for mitigating any skew in the data that could otherwise influence the outcomes of bias analysis. We build on the proposed framework to provide an extensive case study of the bias level of five state-of-the-art detectors in synthetic datasets with 25 controlled facial attributes. While the results confirm that, in general, synthetic face detectors are biased towards the presence/absence of specific facial attributes, our study also sheds light on the origins of the observed bias through the analysis of the correlations with the balancing of facial attributes in the training sets of the detectors, and the analysis of detectors activation maps in image pairs with controlled attribute modifications.
\end{abstract}


\begin{figure}
\centering
\resizebox{0.48\textwidth}{!}{
\begin{tikzpicture}
    \begin{axis}[
        no markers, domain=0.5:0.9, samples=100,
        axis lines*=left, 
        xlabel={Score}, 
        every axis y label/.style={at=(current axis.above origin),anchor=south},
        every axis x label/.style={at=(current axis.right of origin),anchor=west},
        height=6cm, width=9cm,
        enlargelimits=false, clip=false, axis on top,
        xmin=0.4, xmax=0.95,
        ymin=0, ymax=10,
        ytick=\empty, 
        yticklabels=\empty, 
        hide y axis, 
        legend style={at={(0.5,-0.15)}, anchor=north, legend columns=-1, /tikz/every even column/.append style={column sep=0.5cm}}
    ]

    \addplot [very thick,gr1!70!black] {gauss(x,0.7,0.05)};
    \addlegendentry{$f_S(s)^{\text{hair}=\text{blond}}$};
    
    \addplot [very thick,gr0!50!black] {gauss(x,0.75,0.045)};
    \addlegendentry{$f_S(s)^{\text{hair}=\text{black}}$};

    \addplot [very thick,ss!95!ss, dashed] {gauss_gauss(x,0.7,0.05,0.785, 0.045)};
    \addlegendentry{$f_S(s)$};
    
    \addplot [
        draw=none, fill=gr0, fill opacity=0.3
    ] {gauss(x,0.75,0.045)} \closedcycle;
    
    \addplot [
        draw=none, fill=gr1, fill opacity=0.3
    ] {gauss(x,0.7,0.05)} \closedcycle;

    \fill[red!10, opacity=0.5] (0,0) rectangle (axis cs:0.6,9.5);
    \draw[red, very thick, dashed] (axis cs:0.6,0) -- (axis cs:0.6,9.5);
    
    \node[anchor=south west] (A) at (3.65cm, 3.5) {\includegraphics[width=0.8cm]{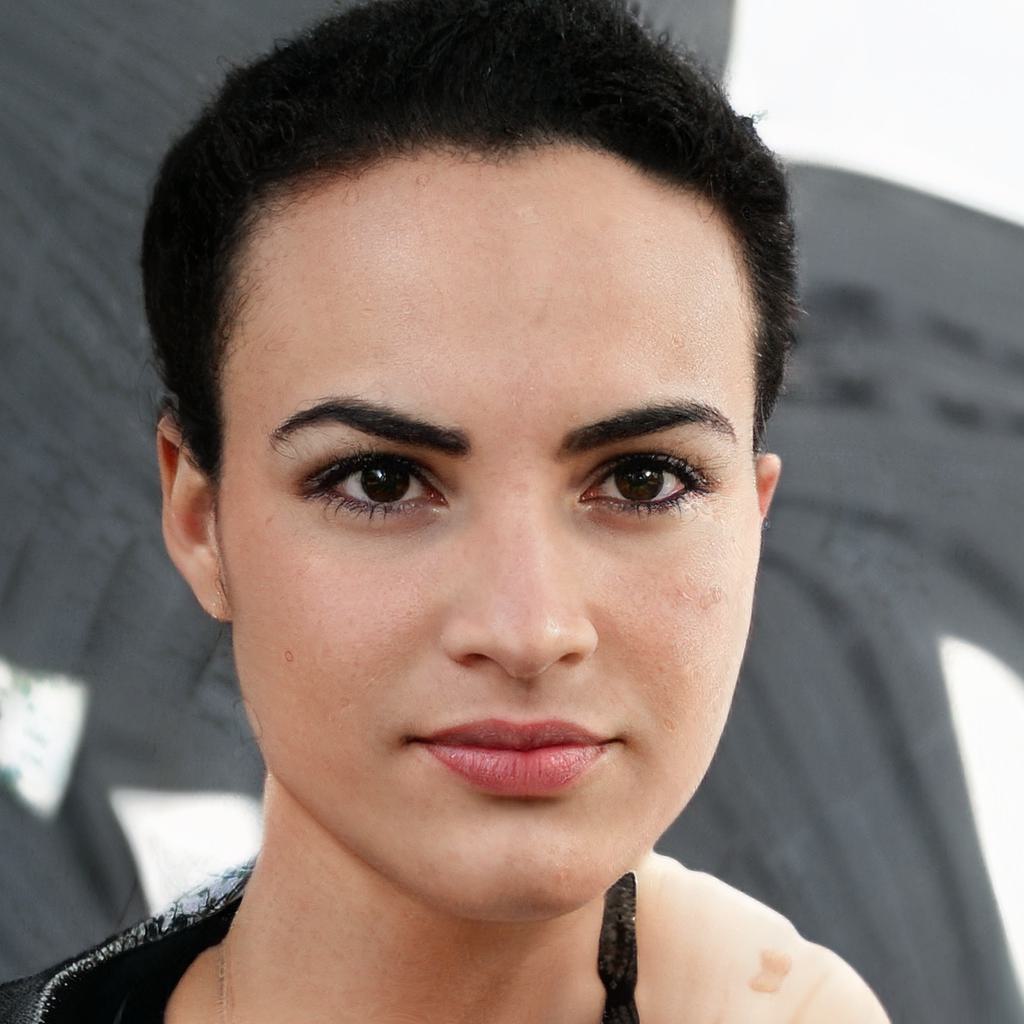}};
    \node[anchor=south west] (B) at (1.55cm, 3.5) {\includegraphics[width=0.8cm]{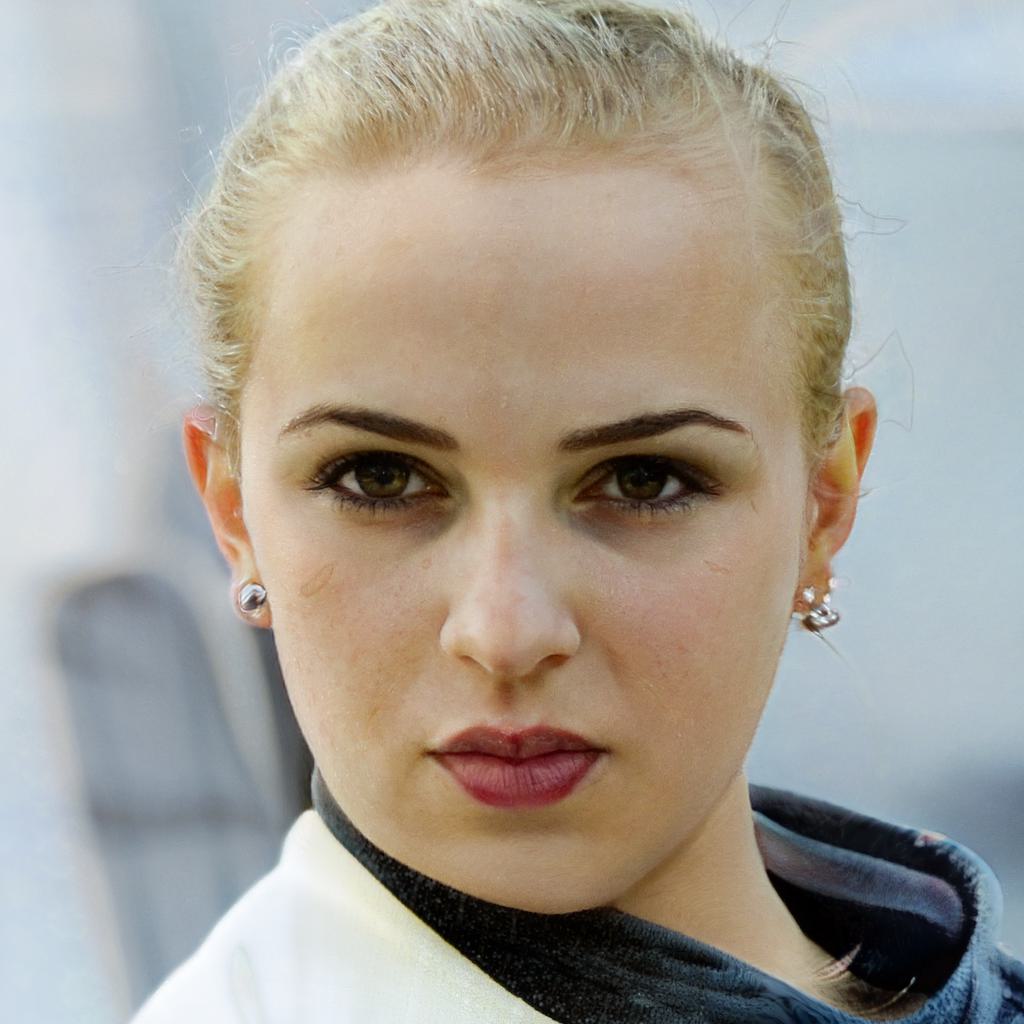}};

    \draw[very thick, ->, shorten >=-3pt, shorten <=-2pt] (A) -- (B);

    \end{axis}
\end{tikzpicture}
}
    \caption{\textbf{Bias in synthetic face detection}. The plot shows the probability density function \( f_S(s) \) of synthetic face detector scores conditioned on hair color attributes (blond vs. black). The distributions differ significantly by attribute, with many samples of blond-haired faces falling below the detection threshold (red dashed line). This results in a higher likelihood of synthetic blond-haired faces being misclassified as real, highlighting a critical bias issue in synthetic face detection systems.}
    \label{fig:teaser}
\end{figure}
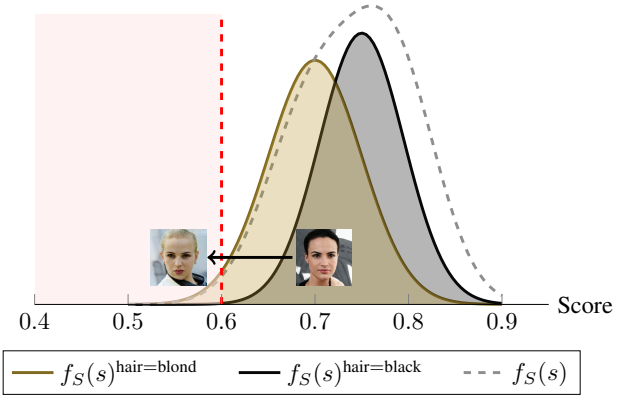
\vspace{-17pt}
\section{Introduction}

Advancements in generative artificial intelligence, particularly through Generative Adversarial Networks (GANs)~\cite{wang2023gan}, have greatly improved the ability to synthesize and manipulate human faces. Techniques such as face swapping, facial reconstruction, and attribute editing now generate highly realistic synthetic content. While these innovations offer valuable applications in media, they also pose significant challenges, particularly concerning security and misinformation. Synthetic face detection systems have been proposed to address these challenges.\\
\noindent\textbf{Motivation.} Despite advancements in detection performance, the fairness and reliability of these systems across diverse demographic groups remain largely unexplored. This contrasts with the extensive studies on bias in facial recognition~\cite{georgopoulos2020investigating}, and given the similarities between traditional face analysis systems and synthetic face detectors, synthetic face detectors are likely to show similar inconsistencies in accuracy across demographics, undermining public trust (Fig.~\ref{fig:teaser}).
Also, the scarcity of balanced datasets with adequate demographic diversity and detailed attribute annotations remains a key issue. While the SDFD dataset~\cite{baltsou2024sdfd} addresses this gap, its limited size of 1,000 images restricts its applicability. \textbf{Research Questions.} To address the identified gap, this work is guided by the following research questions: \textbf{RQ1.} How do the True Positive Rates (TPR) of synthetic face detectors vary when analyzing images with specific facial attributes compared to those without? \textbf{RQ2.} What are the primary sources of bias in synthetic face detection systems? Do these biases originate from the synthetic data generator used for evaluation, the design and training of the detectors, or the composition of their training datasets? \textbf{RQ3.} To what extent do the biases identified in synthetic face detection systems reflect those observed in pristine (non-manipulated) data?
\textbf{Methodology.} To systematically explore these questions, we propose a structured strategy comprising synthetic data generation, bias quantification, statistical validation, and bias source analysis. First, we develop a reproducible methodology for generating synthetic data conditioned on specific facial attributes. A novel metric is introduced to quantify bias by comparing TPRs between groups that differ in only one facial attribute, and statistical validation is carried out using paired t-tests as a bias detection approach. Bias source analysis is performed by visualizing image regions influencing classifier decisions and through the assessment of the correlation between observed biases and the distribution of facial attributes in training datasets.\vspace{5pt} \\
In summary, our contributions are as follows:
\begin{itemize}
\itemsep0em
\item Identification of attribute-specific biases. To the best of our knowledge, this is the first work to investigate the biases of synthetic face detection systems not only in relation to individual facial attributes but also arising from their interactions with other attribute categories.
\item Framework for bias analysis. We introduce a comprehensive framework for analyzing bias in synthetic face detection systems, integrating synthetic data generation, bias quantification metrics, statistical validation, and correlation analysis to provide a robust and systematic evaluation.
\item Synthetic dataset. We present a balanced, labeled synthetic dataset designed specifically for certain facial attribute.
\item Analysis of bias sources. Through systematic evaluation, we identify the primary sources of bias in synthetic face detection systems, including synthetic data generators, training dataset composition, and model design choices.
\end{itemize}
The dataset and framework are publicly available\footnote{\url{github.com/joaocneves/biasdf}} to support reproducibility and foster further research.

\section{Related Work}
\noindent\textbf{Synthetic Face Detection Models}. Synthetic face detection has become a key area of focus with the rise of synthetic content creation~\cite{becattini2023head}. Detection methods are typically categorized into spatial, temporal, and frequency-based approaches~\cite{sandotra2024comprehensive}. Spatial-based methods~\cite{gong2024deepfake,afchar2018mesonet,nguyen2019use,tan2024rethinking} work by detecting visual inconsistencies and pixel-level distortions caused during the creation of synthetic content. Temporal-based methods, primarily applied in videos, detect inconsistencies in motion and coherence across frames. Frequency-based approaches analyze content in the frequency domain, identifying hidden artifacts not visible in the spatial domain. SCnet~\cite{guo2021blind} and Capsule Networks~\cite{sabour2017dynamic} are prominent examples that capture frequency-based features, leveraging device fingerprints and compression artifacts. More recently, transformer-based models such as CORE~\cite{ni2022core}, DFDT~\cite{khormali2022dfdt} and UIA-ViT~\cite{zhuang2020UIA} combine spatial and frequency features for improved synthetic face detection.
\setlength{\textfloatsep}{6pt}
\begin{table}
\resizebox{\columnwidth}{!}{
\begin{tabular}{|l|c|c|c|}
\toprule
Dataset Name & Real/Fake & Unique Combinations & Number of Attributes \\
\midrule
FF++ & Real/Fake & 22529 & 42 \\
CelebDF & Real/Fake & 11740 & 30 \\
DFDC & Real/Fake & 8938 & 43 \\
DFD & Real/Fake & 9944 & 27 \\
\textbf{Ours} & Fake & 46656 & 25 \\
\bottomrule
\end{tabular}
}
\vspace{-1pt}
\caption{Comparison of number of unique combinations.}
\label{tab:dataset_comparison}
\end{table}\\
\noindent\textbf{Synthetic Face Detection Datasets.} Existing synthetic face detection datasets, such as FF++~\cite{roessler2019faceforensicspp}, CelebDF~\cite{li2020celeb}, often exhibit significant shortcomings, including imbalance, limited demographic diversity, incomplete coverage of synthetic face generation techniques, and insufficient attribute annotations~\cite{ju2024improving} (Table \ref{tab:dataset_comparison}).
In this work, to address the urgent need for controlled attributes and a more systematic evaluation of model performance across different groups and conditions, we propose generating datasets that, once specific attributes are defined, systematically include all possible combinations and interactions. This approach ensures a fully balanced dataset, allowing for a more comprehensive and unbiased evaluation of detection models across diverse scenarios.\\
\noindent\textbf{Exploring Bias in the Detection of Synthetic Faces.} Bias analysis in synthetic face detection remains relatively unexplored. Most previous work has focused on evaluating detection algorithms based solely on the presence or absence of a single attribute, usually demographic factors such as gender or race~\cite{ju2024improving}. However, the biases that can emerge from machine learning-based solutions are often more complex and involve interactions among multiple attributes. This hypothesis has been largely confirmed by studies in the literature in various other fields~\cite{mehrabi2021survey}. Previous research has examined bias in synthetic face detection with respect to specific attributes such as race, gender, and skin tone~\cite{trinh2021examination, hazirbas2022towards, nadimpalli2022gbdf, seferbekov2020dfdc, zhou2020kaggle, ntechlab2020deepfake, siyuc2020robustforensics, pan2020dfdc}. Performance disparities have been consistently observed, with models often underperforming on underrepresented groups, particularly individuals with darker skin tones~\cite{hazirbas2022towards}. Even when evaluated on datasets balanced for demographic attributes, models still exhibited biased behavior~\cite{trinh2021examination}. Some efforts attempted to mitigate these issues through dataset rebalancing~\cite{nadimpalli2022gbdf}, which led to modest improvements. However, these approaches were time-consuming, relied heavily on manual annotation, and addressed only isolated attributes, limiting their effectiveness in capturing more complex, intersectional forms of bias.
Biases related to non-demographic attributes have been linked to class imbalances in datasets. Some methods attempted to control for this by fixing one attribute and evaluating performance on balanced subsets~\cite{xu2024analyzing}, but such approaches assume attribute independence and overlook inter-attribute interactions. To address this, our framework introduces a bias risk evaluation metric that jointly assesses performance across multiple attributes, enabling a more comprehensive and nuanced understanding of model bias.

\section{Methodology}

\begin{figure}[t]
\centering
\resizebox{0.45\textwidth}{!}{
\begin{tikzpicture}
    \node[anchor=north west, inner sep=0] at (0,0) {\includegraphics[width=0.1\textwidth]{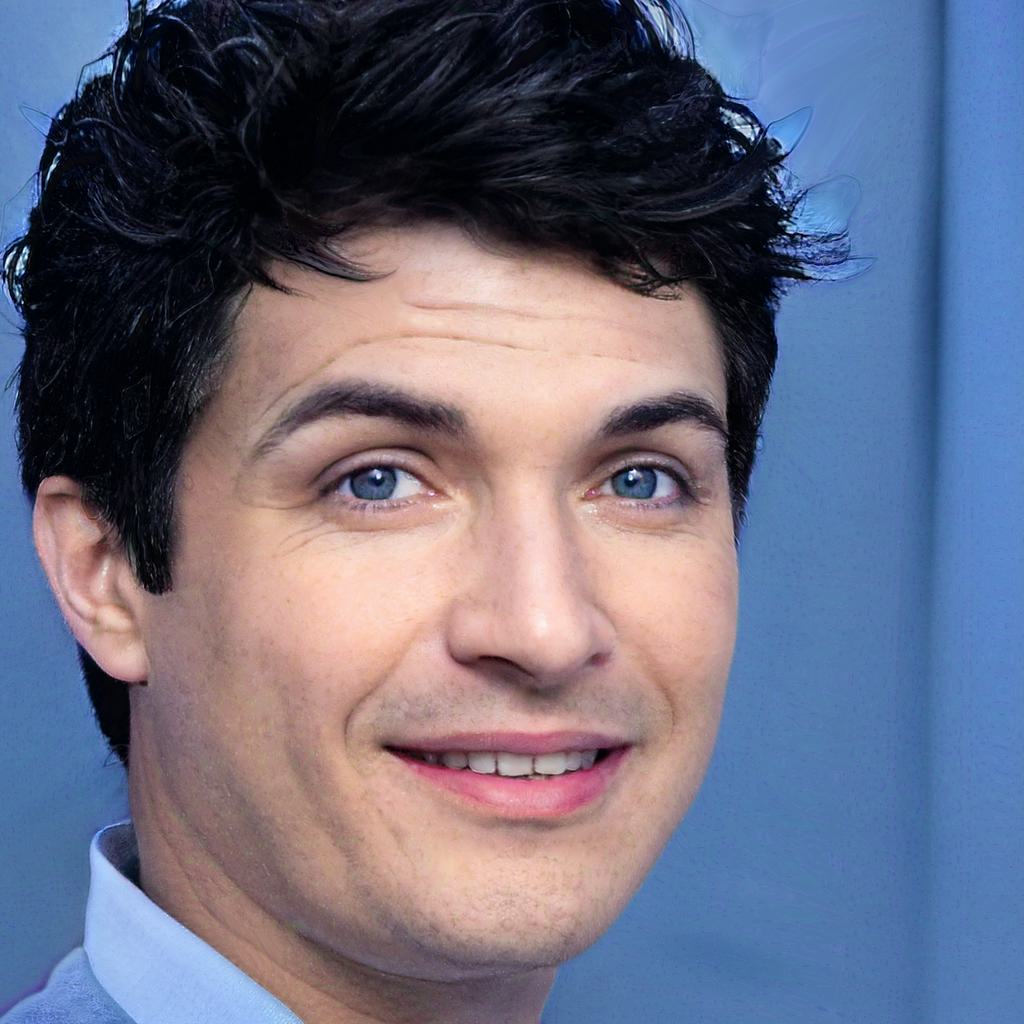}};
    \node[anchor=north west, inner sep=0] at (0.1\textwidth,0) {\includegraphics[width=0.1\textwidth]{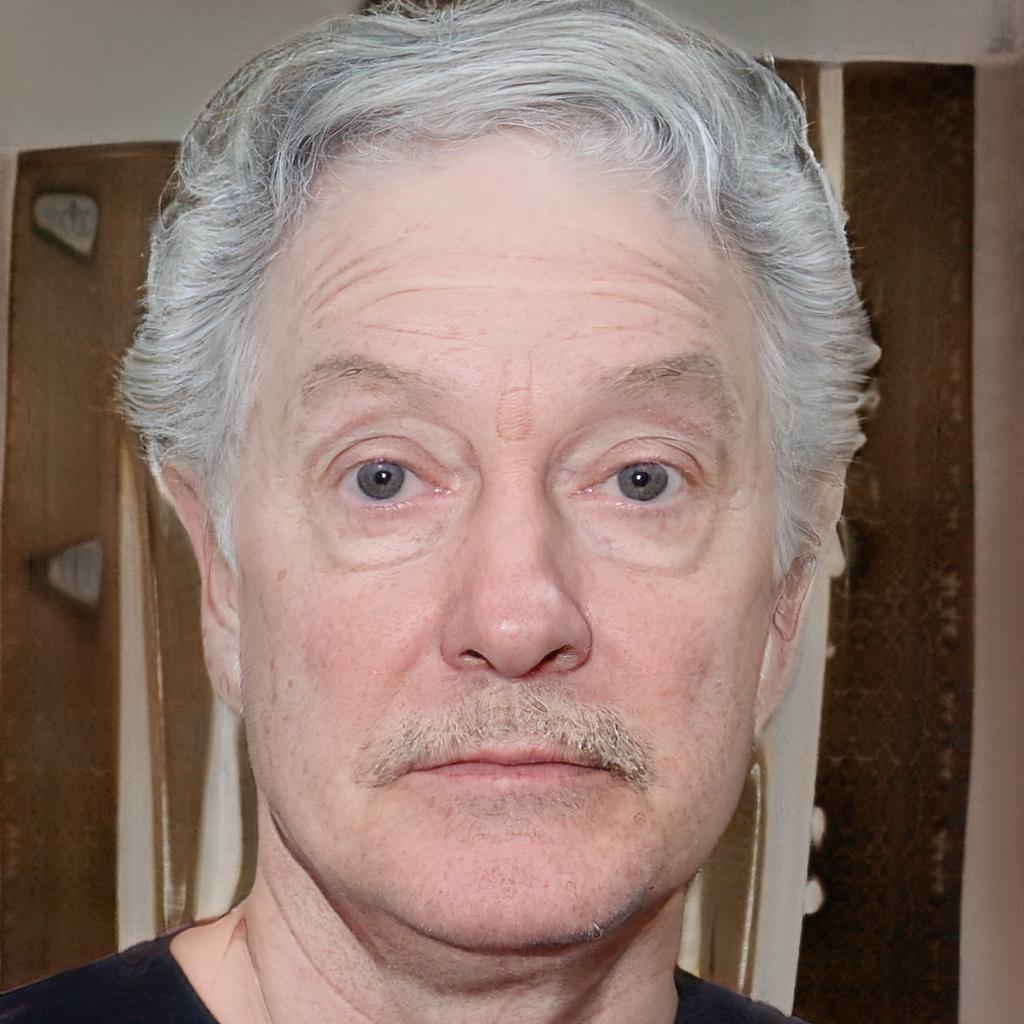}};
    \node[anchor=north west, inner sep=0] at (0.2\textwidth,0) {\includegraphics[width=0.1\textwidth]{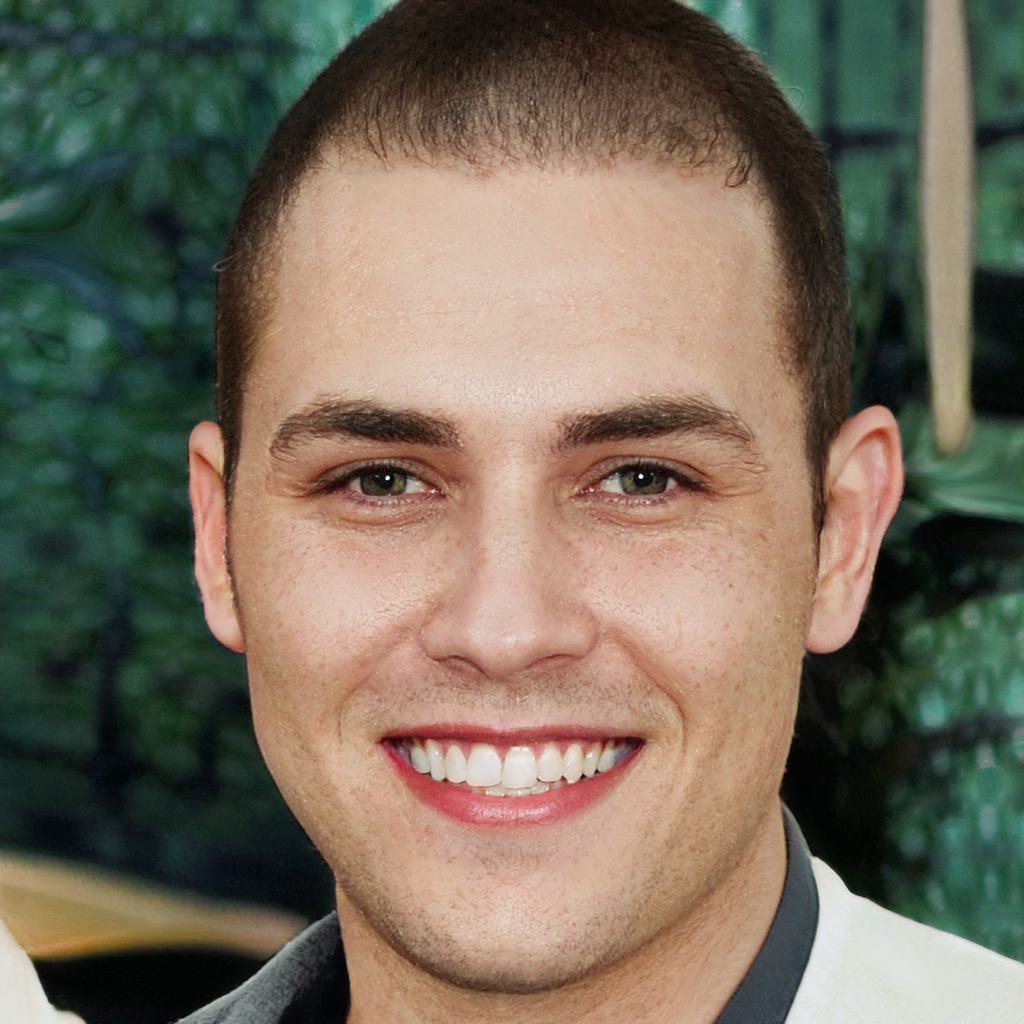}};
    \node[anchor=north west, inner sep=0] at (0.3\textwidth,0) {\includegraphics[width=0.1\textwidth]{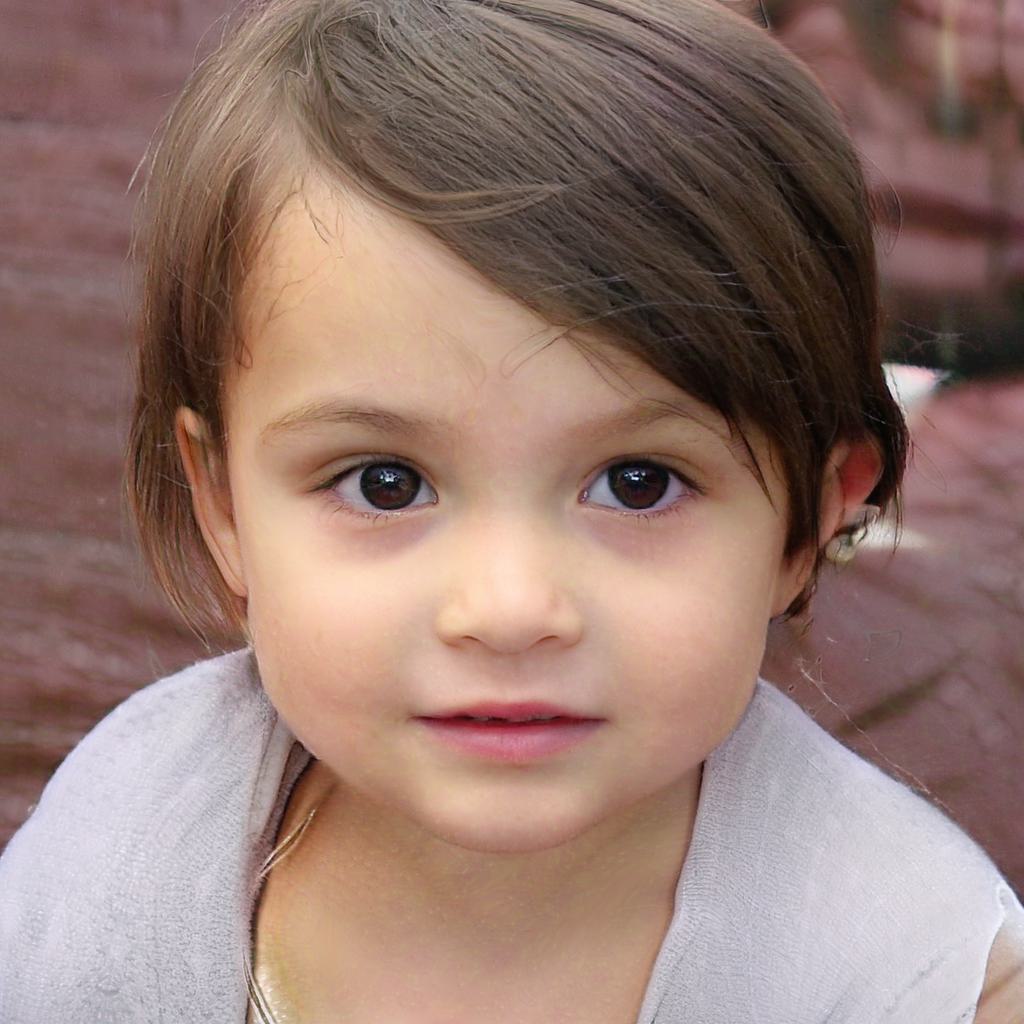}};
    \node[anchor=north west, inner sep=0] at (0.4\textwidth,0) {\includegraphics[width=0.1\textwidth]{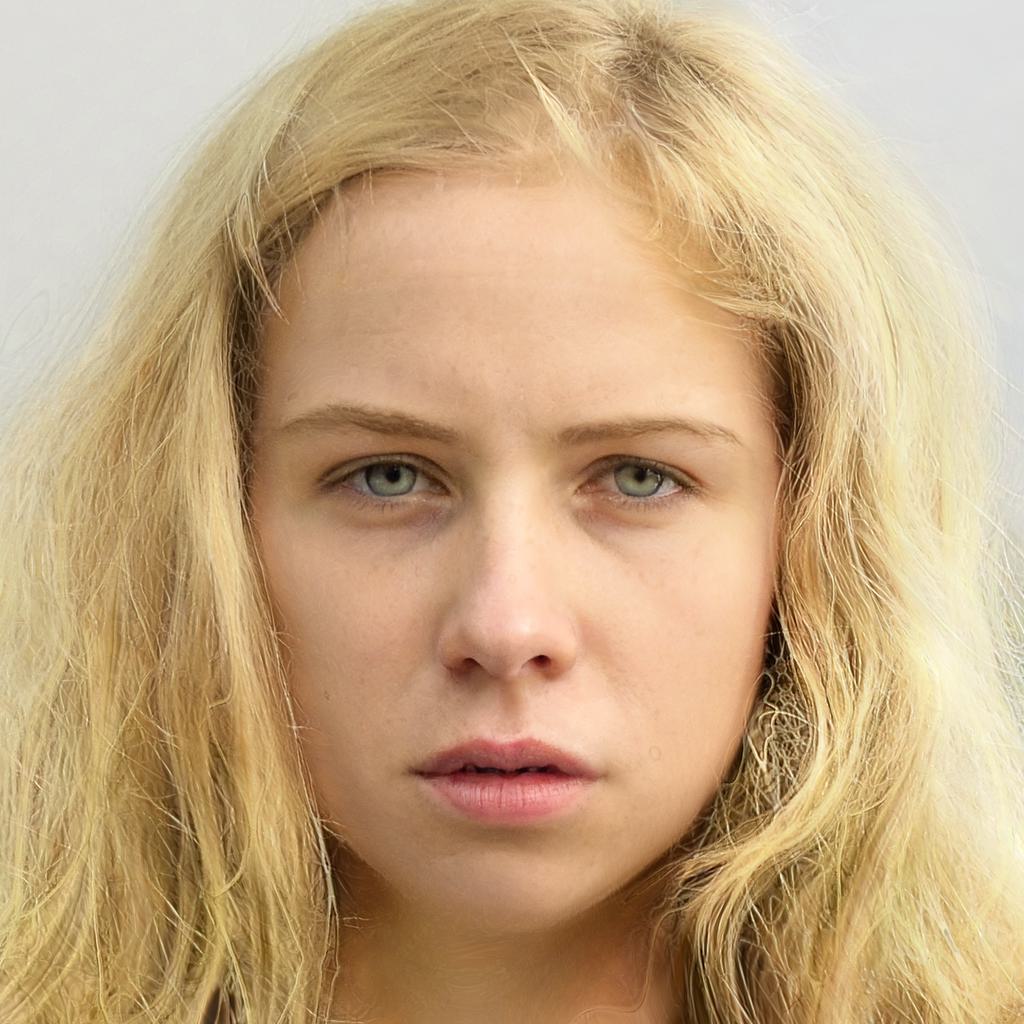}};
    
    \node[anchor=north west, inner sep=0] at (0,-0.1\textwidth) {\includegraphics[width=0.1\textwidth]{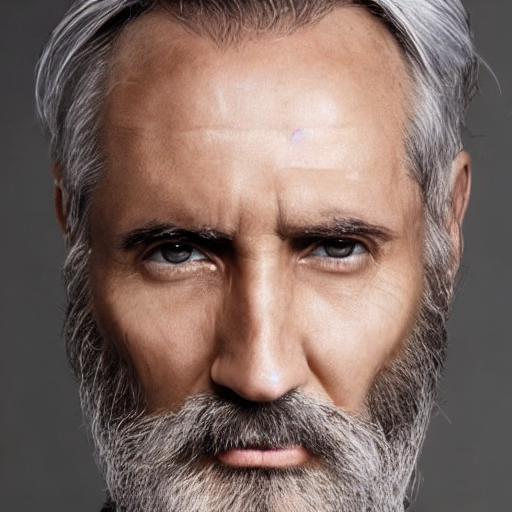}};
    \node[anchor=north west, inner sep=0] at (0.1\textwidth,-0.1\textwidth) {\includegraphics[width=0.1\textwidth]{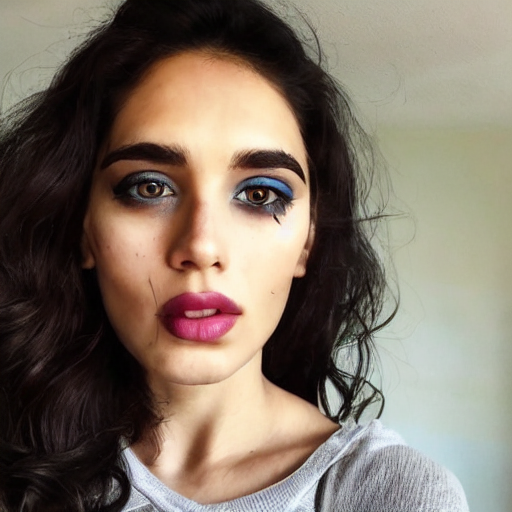}};
    \node[anchor=north west, inner sep=0] at (0.2\textwidth,-0.1\textwidth) {\includegraphics[width=0.1\textwidth]{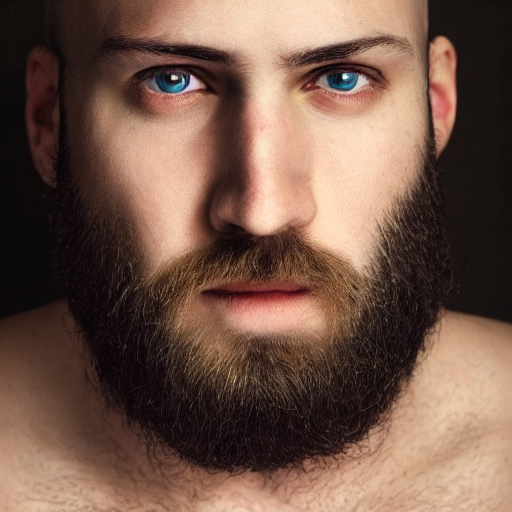}};
    \node[anchor=north west, inner sep=0] at (0.3\textwidth,-0.1\textwidth) {\includegraphics[width=0.1\textwidth]{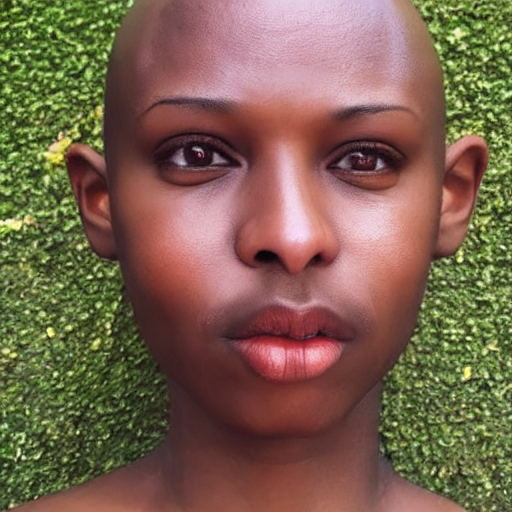}};
    \node[anchor=north west, inner sep=0] at (0.4\textwidth,-0.1\textwidth) {\includegraphics[width=0.1\textwidth]{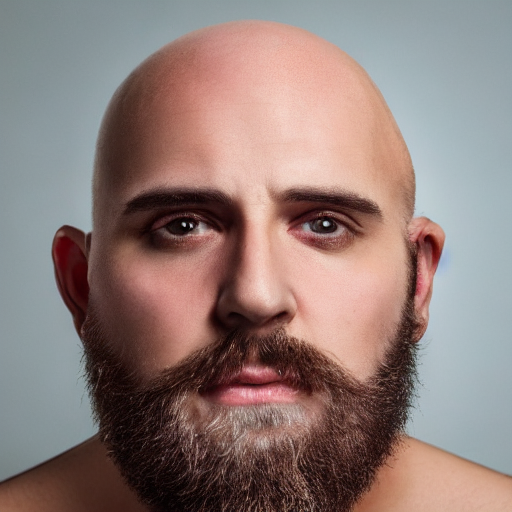}};
\end{tikzpicture}
}
\caption{\textbf{Generated synthetic data}. Samples from the synthetic datasets created by two generators (top row: StyleGAN, bottom: StableDiffusion) representing a diverse range of facial attributes.}
\label{fig:enter-label}
\end{figure}

\subsection{Data Generation}
To conduct a thorough analysis of the impact that different facial attributes have on the performance of synthetic face detectors, we propose generating a synthetic dataset that systematically includes all possible combinations of attribute labels. This approach not only enables a comprehensive evaluation of detection accuracy but also allows for precise control over individual attributes, facilitating the isolation of their specific effects on the model performance. Let $G = \{g_1, g_2, \dots, g_n\}$ be a set of attribute groups, where \( |L_i| \) represents the number of possible attributes for the group \( g_i \in G \). We generate a set of $k$ synthetic images for each  possible label combination across the $n$ groups. The total number of unique combinations for \( G \) is thus given by $|\Omega(G)| = \prod_{g_i \in G }|L_i|$, ensuring full coverage of the attribute space. Figure~\ref{fig:enter-label} illustrates some examples of the synthetic images generated.

\subsection{Bias Risk}\label{subsubsec}
The classical approach to bias assessment typically measures how model predictions differ across demographic groups, which fails to consider the influence of other relevant attributes that may contribute to bias. We propose an alternative measure, where bias comparisons are conducted within subgroups where both demographic and non-demographic attributes are alternately fixed, allowing only the attribute under analysis to vary. This method provides a more comprehensive understanding of bias by accounting for the interactions between multiple attributes.  Building on this concept and utilizing the synthetic data we generated, which allows for precise control and clear understanding of individual attributes, we propose a novel metric, termed bias risk ($brisk$). This metric quantifies bias by calculating the expected variation in True Positive Rates (TPR) across subgroups. These subgroups are defined by fixing the attribute under investigation and systematically varying all other attributes. Formally, let \( A = \{a_1, a_2, \dots, a_n\} \) represent a set of attributes, where \( a_i \) is the attribute under analysis, and let \( A_{-i} = A \setminus \{a_i\} \) denote the set of all other attributes except \( a_i \). Given a classification model \( S \), let \( f_S^{(a_i = x)} \) represent the probability density function of the scores produced by \( S \) for the subgroup where the attribute \( a_i \) takes the value \( x \) (with \( x = 1 \) indicating the presence and \( x = 0 \) indicating the absence of \( a_i \)). The TPR for each subgroup of \( A_{-i} \) is defined as:
\begin{align}
    \text{TPR}(a_i = x, A_{-i}, t) &= \int_t^{\infty} f_S^{(a_i = x)}(s \, | \, A_{-i}, \text{TP}) \, ds, 
    \label{eq:tpr_integral}
\end{align}
where $t$ represents the threshold score at which the classifier operates and $f_S^{(a_i = x)}(s \, | \, A_{-i}, \text{TP})$ represents the probability density function of the classifier's scores $s$, conditioned on the instance being a True Positive (TP) for the fixed attribute $a_i$, and further conditioned on the remaining attributes $A_{-i}$ . To quantify bias within each subgroup, we propose comparing the TPRs for groups with and without the attribute $a_i$:
\begin{equation}
    \begin{split}
    \Delta_{\text{TPR}}(a_i, A_{-i}, t) = &\ \text{TPR}(a_i = 1, A_{-i}, t) \\
    &- \text{TPR}(a_i = 0, A_{-i}, t).
    \end{split}
    \label{eq:tpr_difference}
\end{equation}

To summarize this comparison across all possible subgroups defined by $A_{-i}$, we calculate the average TPR difference across all combinations of $A_{-i}$, denoted by $\Omega(A_{-i})$:
\begin{align}
    \Delta_{\text{TPR}}(a_i, t) &= \frac{1}{|\Omega(A_{-i})|} \sum_{A_{-i} \in \Omega(A_{-i})} \Delta_{\text{TPR}}(a_i, A_{-i}, t).
    \label{eq:average_tpr_difference}
\end{align}

Here, $|\Omega(A_{-i})|$ is the number of possible subgroups formed by the different combinations of attributes in $A_{-i}$.
The function \( \Delta_{\text{TPR}}(a_i, t) \) estimates bias as a function of the operational threshold \( t \), reflecting how the TPR difference between groups changes with different decision boundaries. To provide a comprehensive measure of bias across all possible thresholds, we propose calculating the expected value of \( \Delta_{\text{TPR}}(a_i, t) \) over the entire range of thresholds:

\begin{equation}
    \text{brisk}(a_i) = \int_{0}^{1} \Delta_{\text{TPR}}(a_i, t) \, dt.
    \label{eq:bias_risk}
\end{equation}

This bias risk metric condenses the bias associated with a specific attribute into a single value, capturing variations in performance across different thresholds and subgroups, thereby providing a more robust estimate of the model fairness. Additionally, we introduce the worst-case bias risk,  $\text{brisk}^{\star}(a_i)$ , to assess the most extreme bias that might occur between groups:

\begin{equation}
    \text{brisk}^{\star}(a_i) = max(\Delta_{\text{TPR}}(a_i, t)).
    \label{eq:bias_risk}
\end{equation}

\normalsize

This alternative metric provides a useful insight for applications where even rare occurrences of extreme bias could have substantial consequences.

\begin{table}[t]
\centering
\resizebox{0.49\textwidth}{!}{
\begin{tabular}{|l|l|}
\toprule
\textbf{Attribute Group} & \textbf{Attribute} \\
\midrule
Attractiveness & Attractive, Not Attractive \\
Gender & Man, Woman \\
Age & Child, Young, Old \\
Hair Color & Black Hair, Blonde Hair, Brown Hair, Gray Hair \\
Hair Type & Straight Hair, Wavy Hair, Bald \\
Skin Tone & Black Skin, White Skin \\
Eye Color & Black Eyes, Blue Eyes, Green Eyes \\
Nose Shape & Pointy Nose, Big Nose \\
Face Shape & Oval Face, Round Face, Square Face \\
Facial Hair & Mustache, Beard, Mustache \& Beard \\
Makeup Type & No Makeup, Makeup, Heavy Makeup \\
\bottomrule
\end{tabular}
}
\vspace{-1pt}
\caption{Facial attributes considered in this study organized with respect to the attribute groups.}
\label{tab:attribute_groups}
\end{table}

\subsection{Framework Evaluation Tools}
\label{sec_Framework Evaluation Tools}
\noindent \textbf{Chart of $brisk$ Values.} A chart displaying the $brisk$ values enables direct comparison across different attributes, facilitating the identification of significant biases among various detectors. This visualization allows for a quick assessment of which attributes exhibit pronounced differences in bias, providing critical insights into model performance.\\
\noindent \textbf{Detector Activation Map.} An interpretable visualization tool designed to identify the most relevant image regions in the model’s predictions, thereby providing insights into the model's decision-making process. The heatmaps generated by a saliency-based visualization method illustrate the importance of different regions in relation to the classifier's score. In the visualizations, each image is accompanied by a score that reflects the model's confidence in its prediction. These scores range from 0 to 1, with higher values indicating increased confidence in the classification. The images are organized in a grid format, with each row representing a specific set of attributes where only one attribute is modified while keeping all other attributes constant in the subsequent row. This arrangement allows for the observation of how the model’s focus changes with alterations in a single attribute, offering valuable insights into the consistency of the model's predictions. This methodology enhances our understanding of how different regions of the image impact the model's decisions, facilitating the identification of potential biases and providing direction for further analysis and improvements in model design.\\
\noindent \textbf{Paired t-Test for Bias Detection.} While the $brisk$ metric quantifies the expected difference in TPR when a specific attribute is present, it does not provide a definitive criterion for when a detector should be considered biased towards that attribute. To rigorously determine the presence of bias, we introduce a statistical hypothesis testing approach, specifically employing a paired t-test. Our synthetic dataset is structured to include $k$ samples for each possible combination of attributes. This balanced design ensures that comparisons between groups are not confounded by unequal sample sizes or attribute distributions.
By calculating the average difference in TPR over all thresholds using our metric, we obtain an overall measure of disparity that is straightforward to interpret. The primary objective of our statistical analysis is to test whether this mean difference in TPR between the two groups is statistically significant, which would indicate potential bias. This approach not only quantifies the mean performance difference between groups but also statistically validates whether the observed differences are significant. Formally, we determine the t-test statistic for determining the presence of bias in the attribute $a_i$ by testing whether $\int_{0}^{1} \Delta_{\text{TPR}}(a_i, A_{-i}, t) \, dt$ has zero mean. To control the family-wise error rate across our 250 hypothesis tests (25 attributes × 5 detectors × 2 generators), we apply the Bonferroni correction. Our significance threshold was adjusted to $p < 4 \times 10^{-5}$ $(\alpha = 0.01/250)$, and only results below this corrected value are considered statistically significant.\\
\noindent \textbf{Correlation Analysis.}  This part of our methodology involves two types of correlation analyses. First, it regards the assessment of the relationship between the bias values of a single detector or across multiple detectors and the proportions of samples for each attribute within the training dataset. This provides insights how the distribution of attributes in the dataset may influence the biases observed in the detectors. Second, it evaluates the inter-detector bias correlation, which conveys information about the source of bias, particularly if the bias may be caused by the architectural choices of synthetic face detectors.


\begin{table}[t]
\centering
\footnotesize
\newcommand{\mycol}[1]{#1}
\newcommand{\myheader}[1]{\multicolumn{2}{c|}{#1}}
\resizebox{\linewidth}{!}{
\begin{tabular}{|l|c|c|c|c|c|c|c|c|c|c|}
\hline
 & \myheader{Xcep.} & \myheader{UIA} & \myheader{\makecell{CNN\\LSTM}} & \myheader{NPR} & \myheader{\makecell{Caps.\\Net}} \\
 & S & D & S & D & S & D & S & D & S & D \\
\hline
attractive & \mycol{•} & \mycol{•} & \mycol{•} & \mycol{} & \mycol{•} & \mycol{} & \mycol{} & \mycol{•} & \mycol{} & \mycol{} \\ \hline
man & \mycol{•} & \mycol{} & \mycol{} & \mycol{} & \mycol{•} & \mycol{•} & \mycol{} & \mycol{} & \mycol{•} & \mycol{} \\ \hline
child & \mycol{} & \mycol{} & \mycol{•} & \mycol{} & \mycol{} & \mycol{} & \mycol{} & \mycol{} & \mycol{} & \mycol{} \\ \hline
young & \mycol{•} & \mycol{•} & \mycol{} & \mycol{} & \mycol{•} & \mycol{} & \mycol{} & \mycol{} & \mycol{} & \mycol{} \\ \hline
old & \mycol{•} & \mycol{•} & \mycol{•} & \mycol{} & \mycol{•} & \mycol{} & \mycol{} & \mycol{} & \mycol{} & \mycol{} \\ \hline
black hair & \mycol{•} & \mycol{•} & \mycol{•} & \mycol{} & \mycol{} & \mycol{} & \mycol{} & \mycol{} & \mycol{•} & \mycol{} \\ \hline
blonde hair & \mycol{} & \mycol{•} & \mycol{•} & \mycol{} & \mycol{} & \mycol{•} & \mycol{} & \mycol{•} & \mycol{•} & \mycol{} \\ \hline
brown hair & \mycol{•} & \mycol{} & \mycol{} & \mycol{} & \mycol{•} & \mycol{} & \mycol{} & \mycol{} & \mycol{} & \mycol{} \\ \hline
gray hair & \mycol{•} & \mycol{•} & \mycol{•} & \mycol{} & \mycol{} & \mycol{} & \mycol{} & \mycol{} & \mycol{•} & \mycol{} \\ \hline
straight hair & \mycol{•} & \mycol{} & \mycol{•} & \mycol{} & \mycol{•} & \mycol{} & \mycol{} & \mycol{•} & \mycol{} & \mycol{} \\ \hline
wavy hair & \mycol{•} & \mycol{•} & \mycol{•} & \mycol{} & \mycol{•} & \mycol{} & \mycol{•} & \mycol{•} & \mycol{•} & \mycol{} \\ \hline
bald & \mycol{•} & \mycol{•} & \mycol{•} & \mycol{} & \mycol{•} & \mycol{•} & \mycol{•} & \mycol{•} & \mycol{•} & \mycol{} \\ \hline
white skin & \mycol{•} & \mycol{} & \mycol{} & \mycol{} & \mycol{•} & \mycol{•} & \mycol{•} & \mycol{•} & \mycol{•} & \mycol{} \\ \hline
black eyes & \mycol{•} & \mycol{•} & \mycol{•} & \mycol{} & \mycol{} & \mycol{} & \mycol{} & \mycol{} & \mycol{•} & \mycol{} \\ \hline
blue eyes & \mycol{•} & \mycol{•} & \mycol{} & \mycol{} & \mycol{} & \mycol{} & \mycol{} & \mycol{•} & \mycol{•} & \mycol{} \\ \hline
green eyes & \mycol{} & \mycol{} & \mycol{•} & \mycol{} & \mycol{} & \mycol{} & \mycol{} & \mycol{•} & \mycol{} & \mycol{} \\ \hline
big nose & \mycol{•} & \mycol{•} & \mycol{} & \mycol{} & \mycol{} & \mycol{} & \mycol{} & \mycol{•} & \mycol{•} & \mycol{} \\ \hline
oval face & \mycol{} & \mycol{} & \mycol{•} & \mycol{} & \mycol{} & \mycol{} & \mycol{} & \mycol{} & \mycol{} & \mycol{} \\ \hline
round face & \mycol{} & \mycol{} & \mycol{} & \mycol{} & \mycol{} & \mycol{} & \mycol{} & \mycol{•} & \mycol{} & \mycol{} \\ \hline
square face & \mycol{} & \mycol{} & \mycol{•} & \mycol{} & \mycol{} & \mycol{} & \mycol{} & \mycol{} & \mycol{•} & \mycol{} \\ \hline
mustach & \mycol{•} & \mycol{} & \mycol{•} & \mycol{} & \mycol{} & \mycol{•} & \mycol{} & \mycol{} & \mycol{•} & \mycol{} \\ \hline
beard & \mycol{•} & \mycol{} & \mycol{•} & \mycol{} & \mycol{•} & \mycol{} & \mycol{•} & \mycol{•} & \mycol{} & \mycol{} \\ \hline
no makeup & \mycol{•} & \mycol{•} & \mycol{•} & \mycol{} & \mycol{•} & \mycol{•} & \mycol{•} & \mycol{•} & \mycol{•} & \mycol{} \\ \hline
makeup & \mycol{•} & \mycol{} & \mycol{} & \mycol{} & \mycol{} & \mycol{} & \mycol{} & \mycol{} & \mycol{•} & \mycol{} \\ \hline
heavy makeup & \mycol{•} & \mycol{•} & \mycol{•} & \mycol{} & \mycol{} & \mycol{•} & \mycol{} & \mycol{•} & \mycol{•} & \mycol{} \\ \hline
\end{tabular}
}
\vspace{-1pt}
\caption{Bias detection results for each attribute across detectors evaluated in synthetic data from StyleGAN (S) and Diffusion generators (D) with (•) indicating significant bias after applying a Bonferroni correction for 250 tests (Bonferroni-corrected $p<0.01$).}
\label{tab:bias_detection_paired_ttes}
\end{table}

\section{Experiments}
\subsection{Case Study Setup: Impact of Facial Attributes}
This section details the specific setup for a case study, conducted to demonstrate the application of the framework we have proposed. The study is designed to explore how different facial attributes affect the detection capabilities of synthetic face detection models. It outlines the construction of the synthetic dataset, the selection of synthetic face generators, the deployment of synthetic face detectors, and the metrics used to assess bias.\\\
\noindent\textbf{Dataset.} \label{sec_data} For this study, we utilize two synthetic datasets specifically constructed to enable a comprehensive analysis of biases in synthetic face detection models. In particular, we construct two synthetic datasets by organizing 25 facial attributes into 12 distinct groups, as detailed in Table~\ref{tab:attribute_groups}. The groups follow labels from prior datasets, such as “attractive” from CelebA. For each dataset, all possible combinations of attributes are systematically generated to ensure comprehensive coverage. These combinations condition a synthetic face generator, with the process repeated $k=4$ times to produce multiple samples for each combination.\\

\noindent\textbf{Synthetic Face Generators.} Two state-of-the-art generators are considered for the creation of the synthetic datasets: a) StyleGAN~\cite{stylegan} pre-trained  on the Flickr-Faces-HQ (FFHQ)~\cite{karras2019style} dataset; and b) Stable Diffusion v1.5~\cite{Rombach_2022_CVPR} pretrained on the LAION-5B dataset~\cite{schuhmann2021laion}. These methods were selected for representing the major families of image generation strategies (GAN-based and Diffusion Models).\\
\noindent\textbf{Synthetic Face Detectors.} In our experiments, we used five state-of-the-art synthetic face detection models, namely XceptionNet, CapsuleNet-V2~\cite{nguyen2019use}, LSTM+ResNext model~\cite{abidin2022deepfake}, UIA-VIT~\cite{zhuang2020UIA}, and NPR~\cite{tan2024rethinking}. All models were trained on the FF++ dataset~\cite{roessler2019faceforensicspp}, except for NPR that was trained on the GenImage dataset~\cite{zhu2024genimage}. UIA a Vision Transformer-based model, while the others are CNN-based models. 
\begin{figure}[t]
    \centering
    \includegraphics[trim=0cm 0cm 2cm 1cm, clip, width=0.45\textwidth]{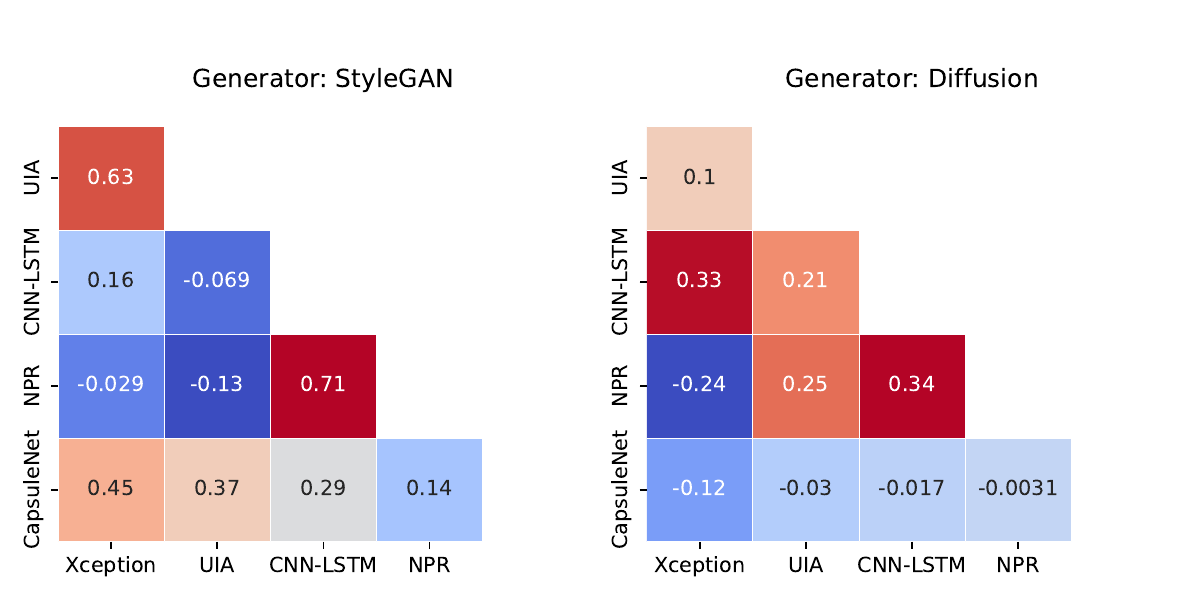}
    \caption{Correlation between the bias values of the different synthetic face detectors.}
    \label{fig:corr_detection}
\end{figure}
\begin{table}[t]
\centering
\resizebox{\linewidth}{!}{
\begin{tabular}{|c|c|c|c|c|c|}
\hline
 & \textbf{Xception} & \textbf{UIA} & \textbf{CNN-LSTM} & \textbf{CapsuleNet} \\ \hline \hline
StyleGAN & 0.03 & -0.16 & 0.63  & -0.09 \\ 
Diffusion & -0.93 & -0.81 & 0.64  & 0.34 \\ \hline
\end{tabular}
}
\vspace{-1pt}
\caption{Correlation between the bias of detectors trained on FF++ and the proportion of samples of each attribute in this set.}
\label{tab:bias_prop_corr}
\end{table}
To assure that the detectors are not biased towards real or fake class, we assessed their accuracy in a balanced set (see supplementary material).\\
\begin{figure*}[ht]
    \centering
    \begin{tikzpicture}
        \node (img1) at (0,0) {\includegraphics[width=0.95\linewidth]{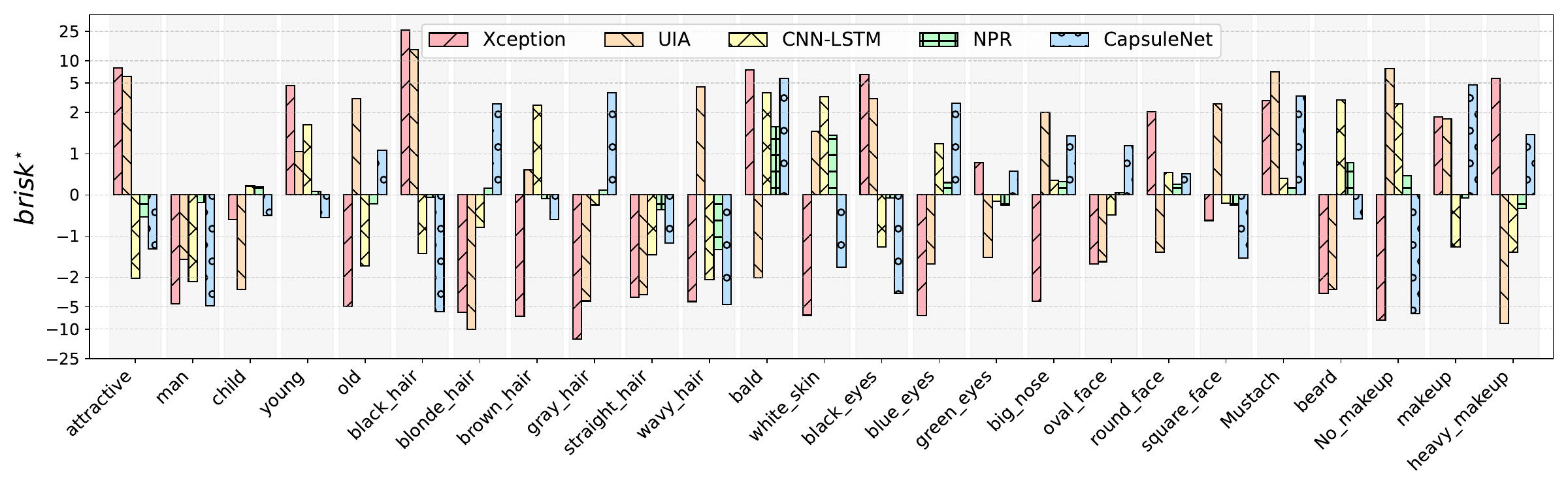}};
        \node[above=of img1, yshift=-1.3cm, text width=0.975\linewidth, align=center] (label1) {Generator: StyleGAN};

        \node (img2) at (0,-5.35) {\includegraphics[width=0.95\linewidth]{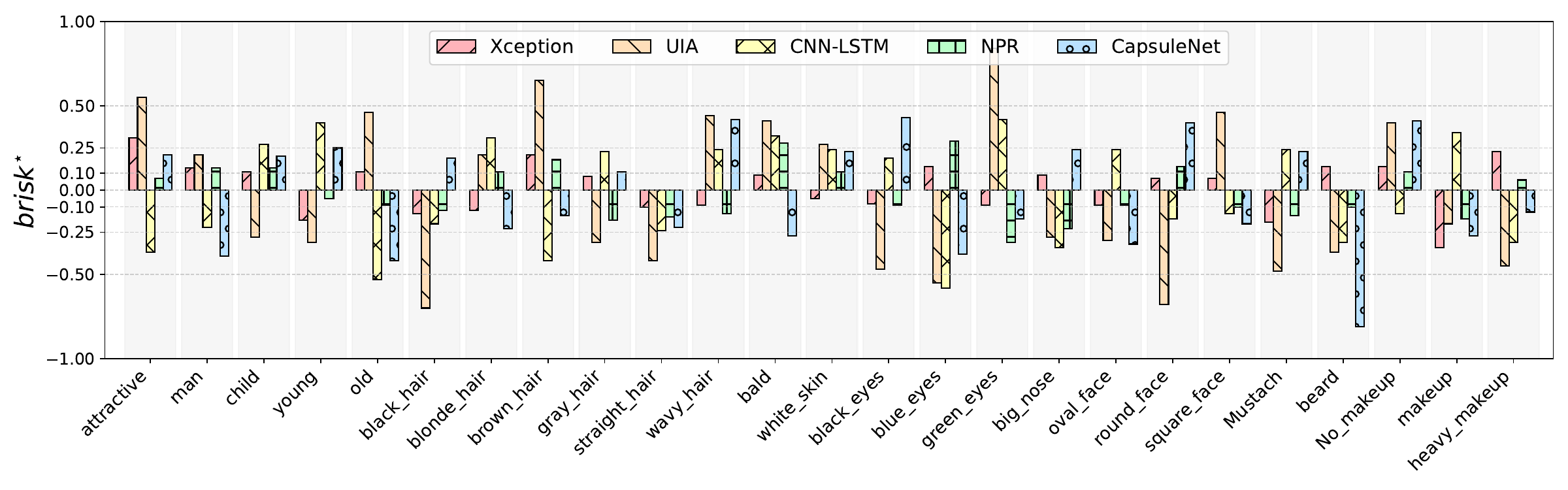}};
        \node[above=of img2, yshift=-1.3cm, text width=0.975\linewidth, align=center] (label2) {Generator: Diffusion Model};
    \end{tikzpicture}
    \vspace{-8pt}
    \caption{\textbf{Bias level of synthetic face detectors along different facial attributes in synthetic data.} The bias level was determined using $\text{brisk}^{\star}$ metric independently for each attribute. Positive values mean that the detector is biased towards the presence of the attribute, implying that samples with this attribute are more likely to be correctly classified as synthetic.}
    \label{fig:brisk_results}
\end{figure*}\noindent\textbf{Metrics.} In our experiments, bias is assessed by comparing the differences in model predictions across various groups while controlling for other attributes using the metric denoted as $brisk$. To complement this analysis, we also utilize the Equal Opportunity Difference (EOD) as an approximation of $brisk$ in scenarios where all subgroups are evenly represented within the dataset. EOD serves as a practical alternative, particularly for existing fake datasets where the complete range of attribute combinations is not available, enabling bias analysis even in the presence of incomplete subgroup representations. Additionally, the methodologies and metrics outlined here utilize the visualization and analysis tools defined in Section \ref{sec_Framework Evaluation Tools}. By integrating these tools, our approach not only quantifies but also visually represents the biases, enabling a clearer understanding of where and how these biases manifest within our models.

\subsection{Results and Discussion}

\noindent\textbf{Bias Assessment.} Our framework was adopted for the assessing the bias level in the five state-of-the-art synthetic face detectors considered in this study. The results of the $\text{brisk}^{\star}$ metric obtained for each detector along 25 facial attributes are depicted in Figure~\ref{fig:brisk_results}. The analysis of the results evidence that some methods are severely affected by bias, as several attributes exceed 5\% in the absolute value of $\text{brisk}^{\star}$. Apart from the absolute bias, it is also worth analysing the sign of the bias, which conveys if the presence of a facial attribute increases (positive bias) or decrease (negative bias) the probability of the detection method to correctly classify the image as synthetic. As an example, the attribute \textit{man} seems to consistently decrease the TPR of the detection method compared to images where this attribute is absent, which in this case corresponds to images of women. One possible reason for this is the unbalanced representation of these two groups in the dataset of the detection methods (42\% man vs 58\% woman in the FF++ dataset). This topic will be further analysed when testing several hypothesis about the origin of bias. Apart from the analysis of the bias level, we report in Table~\ref{tab:bias_detection_paired_ttes} the bias detection results carried out using the paired t-test on the TPR difference between subgroups. After applying a Bonferroni correction to account for the 250 comparisons (corrected $p<0.01$), the results evidenced that several attributes still showed significant bias in specific detectors, confirming that bias in synthetic face detectors is still an open problem.\\
\noindent\textbf{Determining the Origin of the Bias.} Apart from the bias level observed in Figure~\ref{fig:brisk_results}, it is worth noting that only a few attributes exhibit follow a consistent pattern between the value/direction of the biases observed along the different detectors. To provide additional evidence on this, we report the correlation between the bias of the different detectors in Figure~\ref{fig:corr_detection}. In general, the bias of the detectors are weakly or not correlated, suggesting that these bias do not originate on the generators, as each method presents different patterns of bias in the same synthetic data. Considering that most methods were trained in the same dataset (FF++), we inspected the possible relation between the bias of detectors trained on FF++ and the proportion of each attribute in this set. To determine the proportion of samples where each attribute is present, we relied on the annotations provided in~\cite{xu2024analyzing}, and determined the correlation with the bias values of each attribute. These results are provided in Table~\ref{tab:bias_prop_corr}, where no strong positive correlation has been found, suggesting that the unbalanced distribution of the attributes in the training set of the detectors can not be attributed as the source of bias of the methods.
\begin{figure}[t]
    \centering
    \begin{tikzpicture}
        \node[anchor=south west,inner sep=0] (image) at (0,0) {\includegraphics[width=0.95\linewidth]{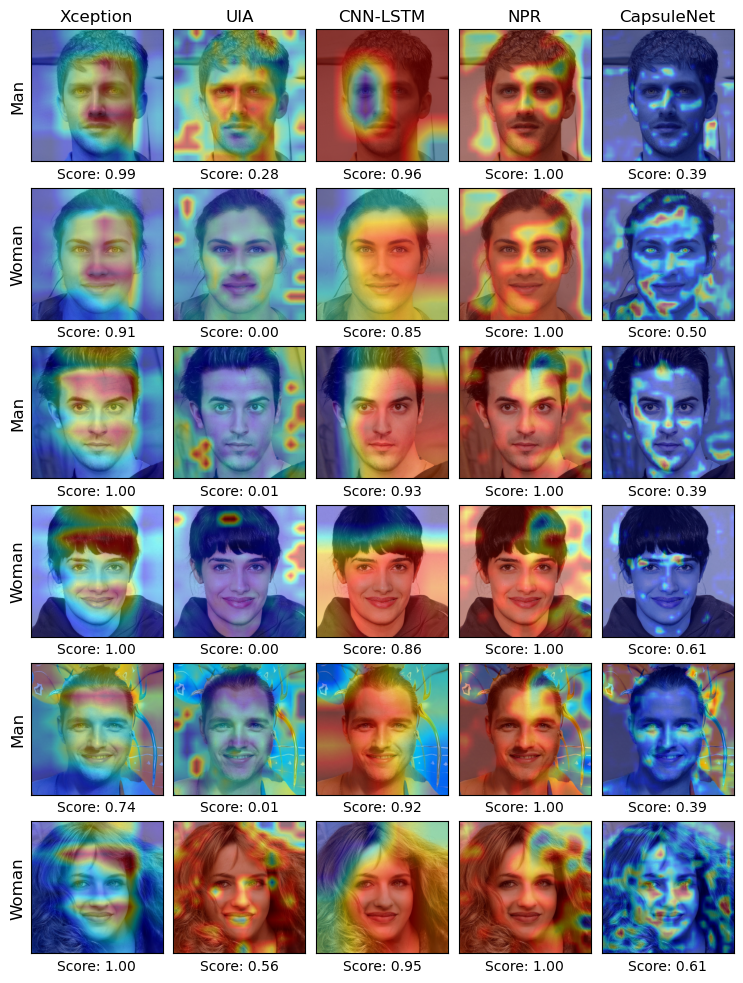}};
        
        \begin{scope}[x={(image.south east)},y={(image.north west)}]
            \draw[rounded corners, line width=2pt, red] (0.2285,0.0058) rectangle (0.2285+0.187,0.0058+0.319);

            \draw[rounded corners, line width=2pt, green] (0.0375,0.653) rectangle (0.0375+0.187,0.653+0.319);

            \draw[rounded corners, line width=2pt, red] (0.805,0.329) rectangle (0.805+0.187,0.329+0.319);

            \draw[rounded corners, line width=2pt, green] (0.613,0.653) rectangle (0.613+0.187,0.653+0.3175);

            \draw[->, thick] (0.025,0.23) -- (0.025,0.14);
            \draw[->, thick] (0.025,0.55) -- (0.025,0.46);
            \draw[->, thick] (0.025,0.87) -- (0.025,0.78); 
        \end{scope}
        
        \node at (4.25,-0.4) { 
            \begin{tikzpicture}
                \draw[red, line width=2pt] (0,0) -- (0.5,0);
                \node[right] at (0.52,0) {{\footnotesize Inconsistent Activation Maps}};
                
                \draw[green, line width=2pt] (4.25,0) -- (4.75,0);
                \node[right] at (4.77,-0) {{\footnotesize Consistent Activation Maps}};
            \end{tikzpicture}
        };
    \end{tikzpicture}
    \vspace{-15pt}
    \caption{\textbf{Synthetic face detectors activation maps.} The activation maps of synthetic face detectors were inferred for pairs of images where only one attribute was changed. The comparison between the activation maps of a pair of images and the respective detection score evidences that in some methods a change in an attribute impacts the regions analysed for determining the classification score, justifying the observed bias with respect to a specific attribute.}
    \label{fig:detectors_activations_gradcam}
\end{figure}
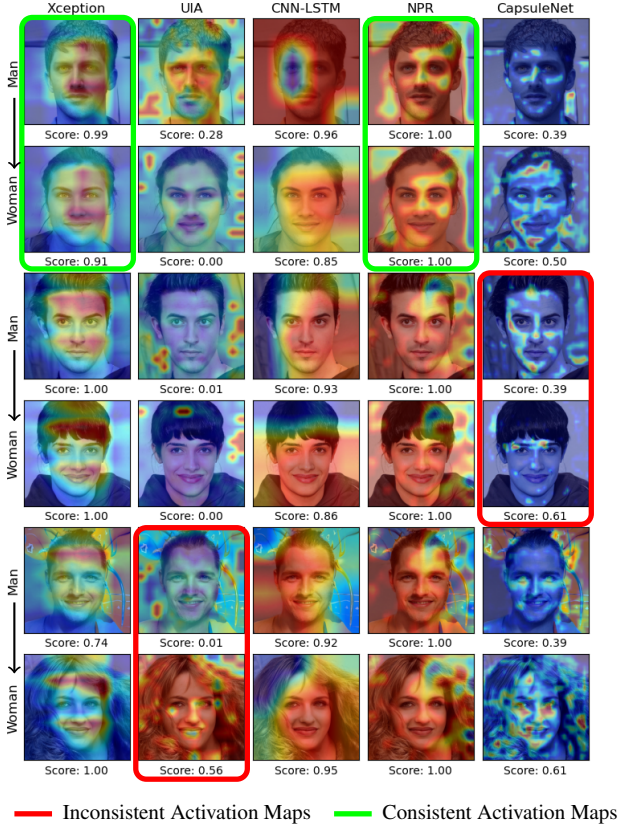
Based on this conclusion, we attempt to verify whether the learning strategy can be related to the bias level of each detector. For this, we inspect the activation maps of face images where only one attribute has been changed. 
The importance of each image region in the classifier score is provided in Figure~\ref{fig:detectors_activations_gradcam}, where the second, fourth and sixth rows have been generated by only inverting the attribute \textit{man}. Additional results on other attributes are provided in supplementary material, showcasing the diverse influence of individual facial features on the classifier's decision-making process. This experiment allows to perceive the consistencies of the detectors to a change in a single attribute, providing insights not only about the origin of the bias but also why some methods suffer more from bias than others. As an example, NPR has the most consistent activation maps when compared with other detectors, and consequently smaller differences between the scores of the groups, and, in turn, less bias. NPR seems to be agnostic to changes in the attribute \textit{man}, while UIA and CapsuleNet present significant differences in the importance regions of the images of the two groups. The results show that architecture and learning strategies may be responsible for the observed bias and not dataset distribution. Some methods, (e.g., UIA and CapsuleNet) seem to be more sensitive to attribute changes, shifting their focus to background when the attribute \textit{man} is modified. This highlights the need for bias-aware learning strategies to enforce detectors to learn invariant features and reduce bias.\\
\noindent\textbf{Importance of Synthetic Data.} To assess the impact of the dataset size in bias estimation, we relied on a simplification of the $brisk$ metric, the EOD metric. EOD measures the TPR difference ($\Delta_{\text{TPR}}(a_i, t)$), but without considering subgroups defined by $A_{-i}$.
\begin{figure}[t]
\raggedright
\resizebox{1.01\linewidth}{!}{
\begin{tikzpicture}
\node at (0, 0) {\includegraphics[width=0.245\textwidth]{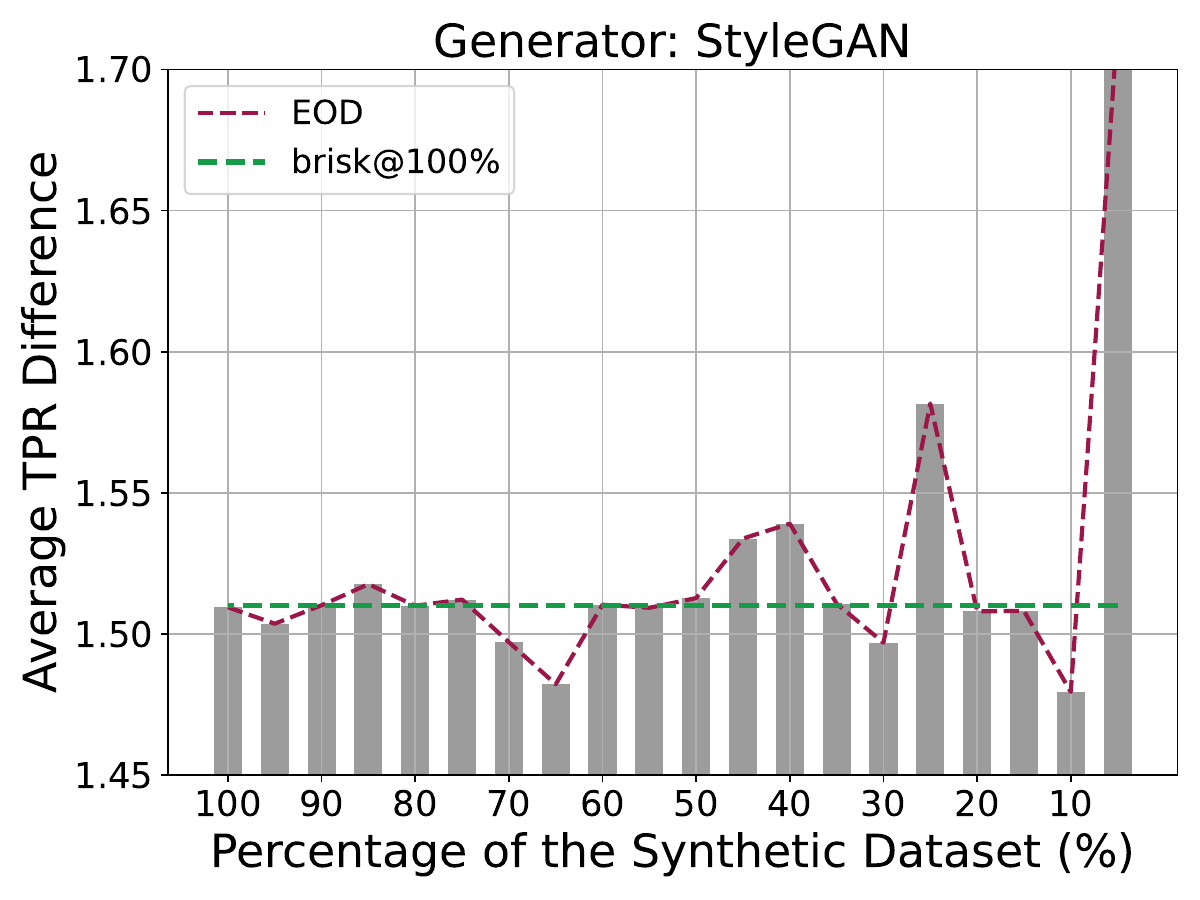}};
\node at (0.25\textwidth, 0) {\includegraphics[width=0.245\textwidth]{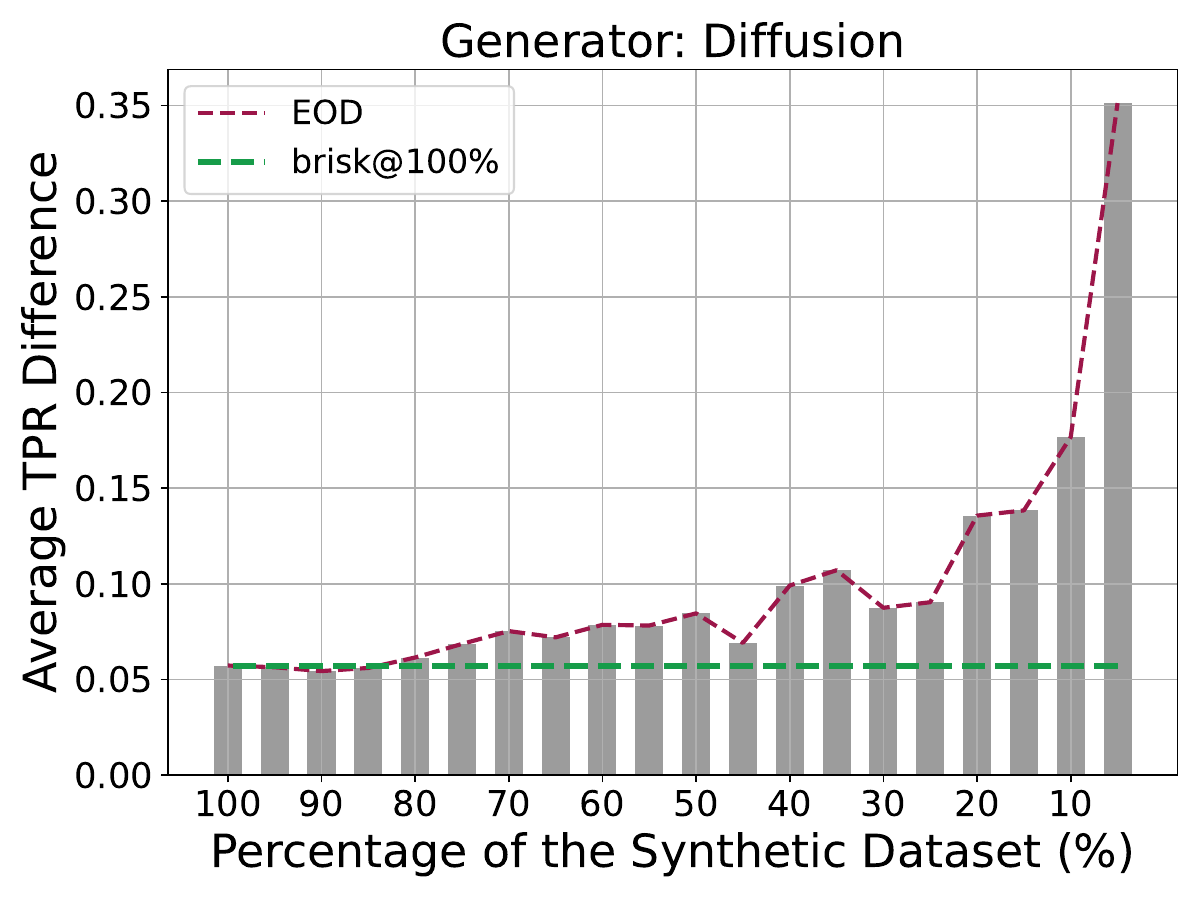}};
\end{tikzpicture}
}
\caption{\textbf{Impact of the representativeness of the subgroups in the estimation of bias.} The average of the EOD metric over the 25 facial attributes was determined using different percentages of our synthtetic dataset. It can be observed that the bias estimation starts diverging from EOD and $brisk$ metrics obtained for a dataset with a balanced number of samples with respect to the different attribute combinations.}
\label{fig:evolution_tpr_diff}
\end{figure}
\begin{table}[t]
    \centering
\resizebox{0.40\textwidth}{!}{
\begin{tabular}{@{}lllcc@{}}
\toprule
\multirow{2}{*}{Gen.} & \multirow{2}{*}{Detector} & \multirow{2}{*}{Attribute} & \multicolumn{2}{c}{\textit{p-value}} \\
\cmidrule(lr){4-5}
  &   &  & Classical & Ours \\
\midrule
\midrule
D & NPR & child & 0.885 & 0.005 \\
S & CapsNet & heavy makeup & 0.720 & 0.000 \\
D & LSTM & brown hair & 0.642 & 0.002 \\
D & LSTM & heavy makeup & 0.220 & 0.000 \\
S & UIA & man & 0.173 & 0.001 \\
D & NPR & old & 0.167 & 0.002 \\
S & CapsNet & square face & 0.152 & 0.000 \\
S & Xception & oval face & 0.139 & 0.007 \\
D & NPR & mustache & 0.120 & 0.005 \\
\bottomrule
\end{tabular}
}
\caption{Comparison between the strategies for deriving a p-value for testing the hypothesis of the TPRs of both $a_i=0$ $a_i=1$ groups not having statistically significant differences.}
\label{tab:paired_ttest_vs_standard_ttest}
\end{table}
Figure~\ref{fig:evolution_tpr_diff} reports the EOD over the different attributes and detectors when using different sample sizes, as well as the $brisk$ metric obtained from the complete synthetic dataset. The results show that the bias estimation is significantly affected when sampling only a small margin of the original synthetic data. We argue that this is caused by the lack of some attribute combinations that impair the accurate estimation of bias. To provide additional evidence on the importance of bias estimation over subgroups when compared to the general strategy of comparing solely the distributions $f_S^{(a_i=1)}(s)$ and $f_S^{(a_i=1)}(s)$, we compared the results of a t-test carried out between the $a_i=1$ and $a_i=0$ groups, and when measuring the differences inside each subgroup sharing the same facial attributes.  Table~\ref{tab:paired_ttest_vs_standard_ttest} reports the top-8 attributes/detectors regarding the difference between the p-values of two approaches, and the comparison between the p-value magnitudes clearly evidences that in all these cases, carrying out a statistical test on the difference between the average TPR of both groups fails to identify bias. In contrast, when considering the average TPR in each subgroup, a statistical significant difference is observed between the two groups, justifying the need for adopting the proposed evaluation methodology for bias estimation.
\begin{figure}[t]
    \centering   \includegraphics[width=0.99\linewidth]{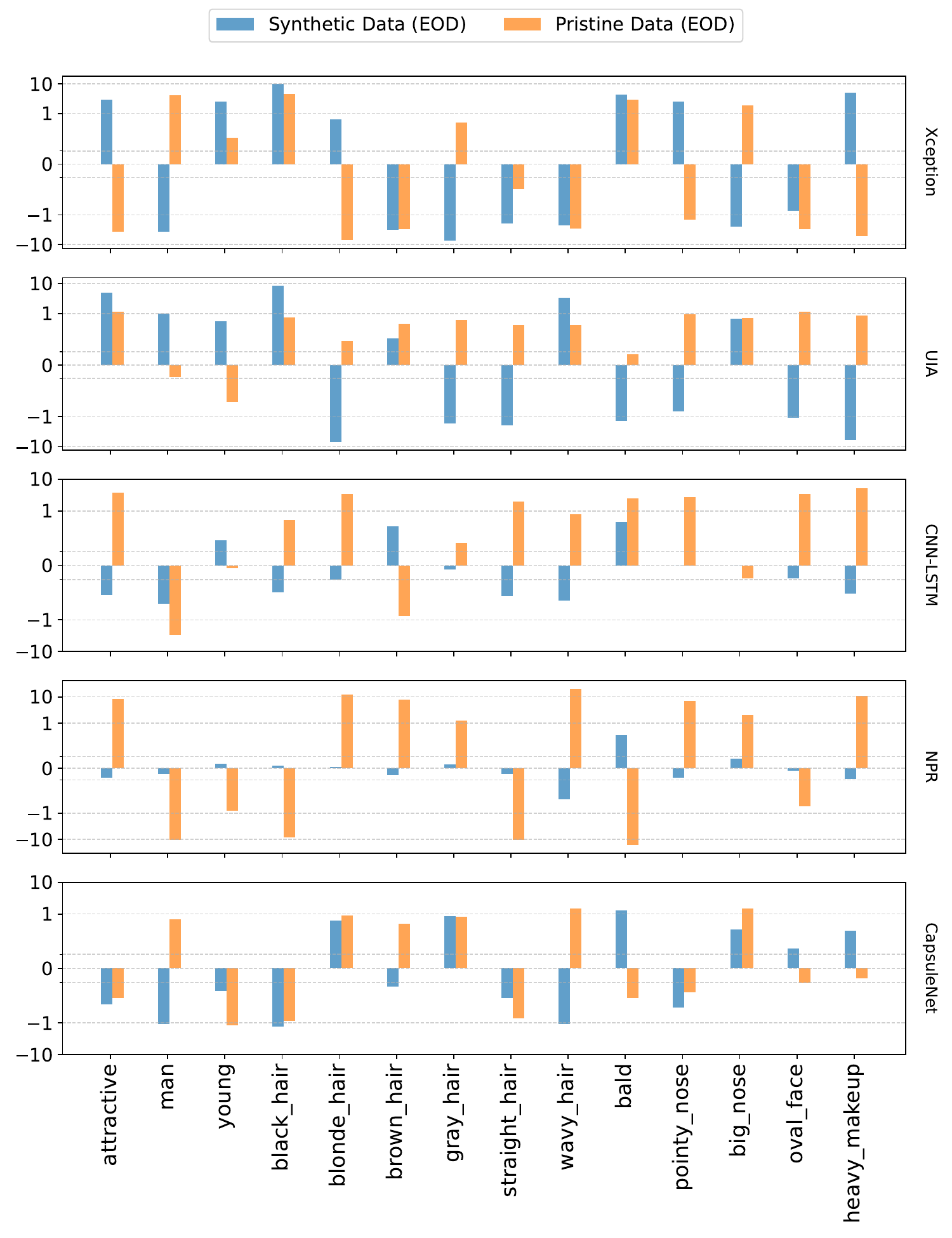}
    \vspace{-2pt}
    \caption{Bias levels of synthetic face detectors in synthetic (blue) and pristine (orange) data, measured across shared facial attributes using the EOD metric.}
    \label{fig:syn_real_bias_comparison1}
\end{figure}

\noindent\textbf{Bias Analysis in Pristine Data.} \label{Bias_Analysis_Pristine}
While the proposed evaluation framework thoroughly analyzes the bias of synthetic face detectors in synthetic data, it does not directly evaluate bias in pristine data due to the impracticality of collecting all possible combinations of attributes. However, to explore the relationship between the biases observed in synthetic and pristine data, we relied on the EOD metric. For this comparison, we selected the CelebA dataset, as it shares several facial attributes with those considered in our study. Figure~\ref{fig:syn_real_bias_comparison1} visually compares the bias levels of different detectors in various facial attributes using synthetic and pristine data. The results demonstrate that biases observed in synthetic data are, in general, also present in pristine data. This discrepancy can be attributed to the varying bias patterns exhibited by different detectors on pristine datasets. Similarly to synthetic data, we observed no strong correlation between a detector's bias level and the proportion of samples with a specific attribute. We hypothesize that this behavior arises from the disparate way classifiers handle attribute information across different prediction score ranges. Specifically, the influence of a facial attribute on bias appears to depend on whether the classifier outputs low scores (pristine data) or high scores (synthetic data). This highlights the distinct dynamics of bias in real-world and synthetic contexts and underscores the value of our synthetic framework for controlled, systematic bias analysis.\\
\noindent\textbf{Limitations.}\label{Limitations}
The proposed bias metric is useful for identifying disparities but should not be used in isolation. A model labeling all inputs as synthetic may appear unbiased despite poor accuracy. We therefore recommend applying the metric only to models that meet a minimum threshold on a holistic performance measure, such as balanced accuracy or precision-recall AUC, ensuring meaningful bias analysis.

\vspace{-4pt}
\section{Conclusion}
\vspace{-2pt}
This study highlights the importance of a systematic framework for uncovering and addressing biases in synthetic face detection systems. Our primary contribution lies in the development of an evaluation strategy that not only identifies the facial attributes most influencing detector decisions but also provides insights into the sources of bias, whether stemming from the detector itself or the training data. While we contribute a synthetic dataset designed for bias analysis, the broader goal of our work is to establish a robust and reproducible methodology for identifying and understanding biases in these systems, particularly those resulting from interactions between facial attributes. This evaluation strategy is dependent on the existence of a dataset encompassing all attribute combinations, which can be obtained using generative methods. Importantly, only the use of a complete set of attributes ensures an accurate estimation of bias. This is evidenced in Figure \ref{fig:evolution_tpr_diff}, where the EOD metric diverges from our proposed metric as the dataset size decreases, highlighting the critical role of comprehensive attribute representation. Through a case study of five state-of-the-art detectors across 25 facial attributes, we observed significant biases in detection accuracy, with TPR differences reaching up to five percentage points in some cases. To identify the sources of these biases, we tested several hypotheses. The lack of strong correlations between biases across different detectors, and between bias levels and facial attribute distributions in the detectors’ training datasets suggest that model architectures and learning strategies play a critical role in introducing biases. This conclusion was further supported by the analysis of activation maps, which highlighted variations in score stability and bias levels across detectors. 
\vspace{-4pt}
\section{Acknowledgments}
\vspace{-6pt}
This work is financed by the project WATERMARK\footnote{WATERMARK project (Watermark-Based Algorithms for Trustworthy Media Authentication and Robust Certification in Public Administration), Project No. 2024.07356.IACDC, supported by “RE-C05-i08.M04 – Support the launch of a program of R\&D projects aimed at the development and implementation of advanced systems in cybersecurity, artificial intelligence, and data science in public administration, as well as a scientific training program,” under the Recovery and Resilience Plan (PRR), as part of the funding agreement signed between the Recovery Portugal Task Force (EMRP) and the Foundation for Science and Technology (FCT).} and supported by UID/04516/NOVA Laboratory for Computer Science and Informatics (NOVA LINCS) with the financial support of FCT.IP.

{\small
\bibliographystyle{ieee}
\bibliography{egbib}
}

\end{document}